\documentclass{amia}
\usepackage{lipsum} %Remove if not needed
\usepackage{graphicx}
\usepackage{booktabs}
\usepackage{multirow}
\usepackage{amsmath}
\usepackage{amssymb}
\usepackage{amsfonts}
\usepackage{xcolor}
\usepackage{ulem}
\usepackage{pifont}
\usepackage{subcaption} 
\usepackage{caption}  
\usepackage{makecell}
\usepackage{fancyhdr}
\fancypagestyle{firstpagefoot}{%
  \fancyhf{}% 清空原有的 header/footer
  \fancyfoot[L]{\small Already accepted into the AMIA 2025 proceedings.}

}

\begin{document}

% max: 10 pages
\title{EnECG: Efficient Ensemble Learning for Electrocardiogram Multi-task Foundation Model}

\author{Yuhao Xu$^1$, Xiaoda Wang$^1$, Jiaying Lu, PhD$^2$, Sirui Ding, PhD$^3$, Defu Cao$^4$, \\Huaxiu Yao, PhD$^5$, Yan Liu, PhD$^4$, Xiao Hu, PhD$^2$, Carl Yang, PhD$^{1, 2}$ }

\institutes{
    {$^1$ Department of Computer Science, Emory University, Atlanta, GA, USA}\\
    {$^2$Center for Data Science, School of Nursing, Emory University, Atlanta, GA, USA}\\
    {$^3$ Bakar Computational Health Sciences Institute, UCSF, San Francisco, CA, USA}\\
    {$^4$Department of Computer Science, USC, Los Angeles, CA, USA}\\
    {$^5$Department of Computer Science, UNC, Chapel Hill, NC, USA}
}

\newcommand{\Xiaoda}[1]{\textcolor{orange}{\#Xiaoda:~#1\#}}

\maketitle
\thispagestyle{firstpagefoot}

\section*{Abstract}

\textit{Electrocardiogram (ECG) analysis plays a vital role in the early detection, monitoring, and management of various cardiovascular conditions. 
% Challenge
While existing models have achieved notable success in ECG interpretation, they fail to leverage the interrelated nature of various cardiac abnormalities. Conversely, developing a specific model capable of extracting all relevant features for multiple ECG tasks remains a significant challenge. Large-scale foundation models, though powerful, are not typically pretrained on ECG data, making full re-training or fine-tuning computationally expensive.
% Contribution
To address these challenges, we propose EnECG~(Mixture of Experts-based \underline{En}semble Learning for \underline{ECG} Multi-tasks), an ensemble-based framework that integrates multiple specialized foundation models, each excelling in different aspects of ECG interpretation. Instead of relying on a single model or single task, EnECG leverages the strengths of multiple specialized models to tackle a variety of ECG-based tasks. To mitigate the high computational cost of full re-training or fine-tuning, we introduce a lightweight adaptation strategy: attaching dedicated output layers to each foundation model and applying Low-Rank Adaptation (LoRA) only to these newly added parameters. We then adopt a Mixture of Experts (MoE) mechanism to learn ensemble weights, effectively combining the complementary expertise of individual models. 
% Experiments
Our experimental results demonstrate that by minimizing the scope of fine-tuning, EnECG can help reduce computational and memory costs while maintaining the strong representational power of foundation models. This framework not only enhances feature extraction and predictive performance but also ensures practical efficiency for real-world clinical applications. The code is available at \url{https://github.com/yuhaoxu99/EnECG.git}}
%}}

\section*{Introduction}

Electrocardiography (ECG) is a quick, painless test that measures the electrical activity of the heart, which is widely used for diagnosing and monitoring various heart conditions~\cite{lu2025cardiac}. Due to its non-invasive nature and the relative ease with which ECG signals can be recorded, ECG analysis remains at the forefront of early detection and management of cardiovascular diseases~\cite{alghatrif2012brief}. Over the past decades, numerous methods~\cite{biel2001ecg, xia2013cloud} have been proposed to extract clinically relevant features from ECG signals. Additionally, the development of large-scale annotated databases~\cite{moody2001impact, goldberger2000physiobank} has facilitated robust benchmarking and spurred advances in ECG analysis.

Despite these achievements, much of the existing work has addressed specific medical task~\cite{adnane2009development, wan2024meit}. This specific medical task focus can overlook the intricate interplay among multiple cardiac abnormalities, leading to redundant processing steps and potentially missing valuable cross-task information. Given that many cardiac conditions, like arrhythmias, ischemia, and myocardial infarction can coexist. If we could not only assess a patient’s cardiac condition but also estimate their blood potassium levels from the same ECG data, we would eliminate the need for blood draws to measure potassium. This strategy could lower healthcare costs while also improving patient comfort. So, there is a strong rationale for more comprehensive approaches that analyze ECG signals in a multi-task manner.

Multi-task learning (MTL) offers a promising framework by concurrently performing several clinically relevant tasks~\cite{caruana1997multitask} simultaneously. This approach capitalizes on shared representations: the feature extraction that benefits one task can also enhance the performance of others. In practice, MTL can lead to better generalization, fewer computational redundancies, and more streamlined clinical workflows. By integrating diverse predictive and diagnostic capabilities into a single system, multi-task ECG models have the potential to offer a broader and more accurate picture of cardiac health, ultimately supporting improved risk stratification, timely clinical interventions, and better patient outcomes.

\begin{figure}[t]
    \centering
    % 子图 (a)
    \begin{subfigure}[b]{0.3\textwidth}
        \centering
        \includegraphics[width=\textwidth]{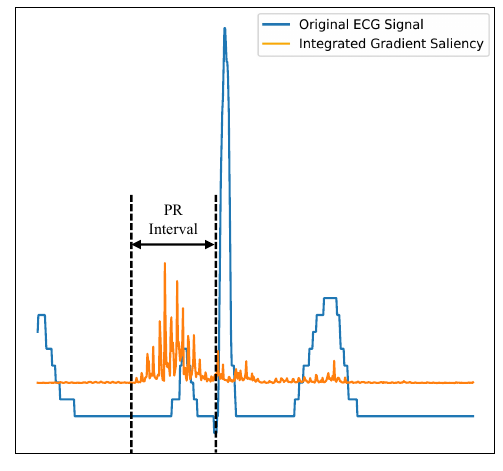}
        \caption{MOMENT}
        \label{fig:a}
    \end{subfigure}
    \hfill
    % 子图 (b)
    \begin{subfigure}[b]{0.3\textwidth}
        \centering
        \includegraphics[width=\textwidth]{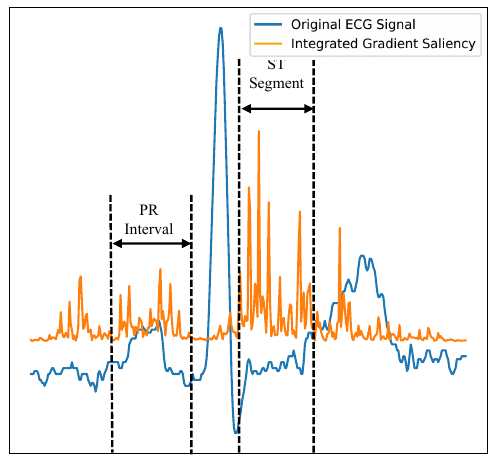}
        \caption{TEMPO}
        \label{fig:b}
    \end{subfigure}
    \hfill
    % 子图 (c)
    \begin{subfigure}[b]{0.3\textwidth}
        \centering
        \includegraphics[width=\textwidth]{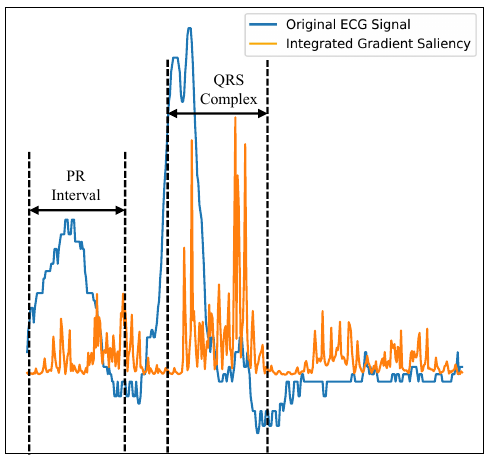}
        \caption{ECG-FM}
        \label{fig:c}
    \end{subfigure}
    \caption{Integrated Gradient Saliency Maps for the RR interval estimation task. As shown, MOMENT primarily focuses on the PR interval, TEMPO draws attention to both the PR interval and ST segment, and ECG-FM emphasizes the PR interval and QRS complex.}
    \label{saliency}
\end{figure}

In addition to the need for analyzing multiple clinical tasks within a single framework, an effective multi-task ECG system requires the extraction of diverse, often complementary features from the signal. Different models may capture different features of specific wave segments, and different wave segments may be more suitable for different clinical tasks. For instance, RR intervals (the time between consecutive heartbeats, reflecting heart rate and variability), QT intervals (the time from ventricular depolarization onset to repolarization completion, representing ventricular action potential duration), and ST segment slope (the inclination of the segment between ventricular depolarization end and repolarization start, indicating potential myocardial ischemia or injury) and morphological features may be helpful for patient age estimation~\cite{attia2019age}, whereas the duration, amplitude, and area of P waves, as well as the QRS complex interval could be advantageous for sex prediction~\cite{siegersma2022deep}. This specialized nature means that no single model architecture can consistently and optimally extract all the important features necessary for performing multiple tasks simultaneously~\cite{caruana1997multitask}. To validate the above findings and gain interpretable insights into the model's decision-making process, we employ the Integrated Gradient Saliency Map~\cite{sundararajan2017axiomatic}. This technique attributes the model's prediction to specific input features as shown in Figure~\ref{saliency}, quantitatively highlighting which parts of the input most significantly contributed to the result. This visualization allows us to assess whether the model relies on semantically relevant features, thereby validating the plausibility and robustness of our findings.

% \begin{figure}[htbp]
%     \centering
%     \includegraphics[width=0.8\textwidth]{AMIA-Latex-Template-master/saliency.pdf}
%     \caption{Integrated Gradient Saliency Map. As shown, MOMENT primarily focuses on the PR Interval, TEMPO draws attention to both the PR interval and ST Segment, and ECG-FM emphasizes the PR Interval and QRS Complex.
%     %\Xiaoda{It would be better to label which part corresponds to the PR interval, ST segment, and QT interval in the original signal.}
%     }
%     \label{saliency}
% \end{figure}

To overcome this limitation, ensemble learning~\cite{dietterich2000ensemble} has emerged as a promising strategy. Rather than relying on a single model, ensemble methods integrate the outputs or feature representations of multiple specialized models, each contributing its unique strengths. By leveraging these complementary capabilities, ensemble-based multi-task ECG models can achieve enhanced robustness and higher overall performance compared to individual models. However, traditional ensemble methods often employ a fixed-weighted average of multiple models’ outputs or a majority voting mechanism, applying the same weighting strategy across all samples. This static approach may not fully exploit the diverse strengths of different models in varying scenarios. To address this, we introduce a Mixture-of-Experts (MoE)~\cite{jacobs1991adaptive, shazeer2017outrageously} gating network, which dynamically assigns weights to each expert model based on the input features. This allows the ensemble to adaptively emphasize the most relevant expert for each specific case, resulting in a more flexible and context-aware fusion of ECG models.
%This approach ultimately provides clinicians with a more comprehensive tool for completing clinical multi-tasks~\cite{pan1985real, polikar2006ensemble}.

Building on the aforementioned advantages of ensemble learning, this work aims to leverage multiple existing foundation models to enhance feature extraction capabilities in multi-task ECG analysis. Foundation models, large-scale pretrained networks, have proven to be highly effective in various domains, but they are rarely pretrained on ECG data and tasks. Consequently, fully retraining or fine-tuning such models can become prohibitively resource intensive, and because hospitals have limited training resources, fully retraining or fine-tuning these models can be challenging. To overcome these limitations, this paper introduces an innovative solution: we add a Feedfoward layer to each foundation model and adopt Low-Rank Adaptation (LoRA)~\cite{hu2021lora} to fine-tune it. By focusing on a smaller set of parameters and employing LoRA’s parameter-efficient training strategy, we can significantly reduce the computational overhead and GPU memory, while still achieving state-of-the-art performance across multiple ECG tasks.

We summarize our contributions: \textbf{(i)} We propose EnECG, an ensemble-based efficient model designed to handle multiple ECG tasks in a unified framework. \textbf{(ii)} We conduct a comprehensive evaluation by selecting five diverse ECG tasks under a multi-task setting, demonstrating favorable performance across all of them. \textbf{(iii)} We open-sourced our code to facilitate reproducibility and foster further research in the community.

\section*{Related Work}

% Jiaying suggested outline: this subsection to introduce background of the multi-tasks used in the experiments.
% early efforts focus on analyzing the ECG itself, including QRS detection, ST segmentation. 
% Later researchers explored arrthymia detection.
% Moreover, researchers find that ECG can be human biometrics, therefore can be used to estimate health parameters such as sex and age.
% Some novel application for ECG include utilizing it as an non-invaisive measurement to estimate traditionall invasive blood lab result. In our paper, we use Potassium.
\uline{\textit{ECG based Clinical Applications}}. Early efforts in automated ECG analysis predominantly focused on single-task objectives, such as arrhythmia classification, QRS detection, or waveform segmentation supported by notable datasets, MIT-BIH Arrhythmia Database~\cite{goldberger2000physiobank} and classical algorithms the QRS detection method proposed by Pan and Tompkins~\cite{caruana1997multitask}. However, cardiovascular pathologies often manifest with interlinked abnormalities, necessitating analytical models that can handle multiple tasks concurrently.

Recent work has explored deep learning strategies for multi-task ECG analysis, leveraging shared representations to perform tasks like rhythm classification, beat segmentation, and clinical parameter estimation within a single framework. For instance, Yao et al.~\cite{polikar2006ensemble} employed a multi-head convolutional neural network to jointly classify arrhythmic beats and estimate heart rate variability metrics. Their model demonstrated that features extracted for one task can enhance performance on another task. Similarly, Chang et al.~\cite{hu2021lora} proposed a multi-task learning architecture based on attention mechanisms to detect multiple types of cardiac abnormalities simultaneously, highlighting that MTL can outperform single-task models when tasks are related. In another study, Zhao et al.~\cite{pan1985real} designed an MTL system for arrhythmia detection and ST-segment deviation analysis, emphasizing the value of parameter sharing to improve generalization and reduce the risk of overfitting in clinical settings.

Despite these advances, a significant challenge persists: no single model architecture uniformly excels at extracting all relevant ECG features required for diverse tasks. As a result, ensemble learning has garnered attention for multi-task ECG analysis. By combining multiple specialized models each trained to capture distinct signal properties an ensemble approach can achieve a more comprehensive characterization of cardiac activity~\cite{moody2001impact}. Nevertheless, a major bottleneck is that many large-scale “foundation models” are trained on general-domain data rather than ECG signals, making them computationally expensive to fully retrain or fine-tune for multi-task ECG applications. Approaches that selectively fine-tune only parts of these large models represent a promising direction for resource-efficient and high-performing multi-task ECG analysis.

\uline{\textit{Ensemble Learning}}. Ensemble learning has long been recognized as a potent strategy to boost predictive performance by combining multiple learners, each with its unique strengths and weaknesses~\cite{breiman1996bagging}. Common ensemble techniques include bagging, boosting, and stacking. For instance, Dietterich~\cite{dietterich2000ensemble} showcased how bagging and boosting can reduce variance and bias in decision trees, while Polikar~\cite{polikar2006ensemble} provided a comprehensive overview of ensemble-based decision-making systems, highlighting their robustness to noise and model uncertainty. These foundational works established the theoretical underpinnings and practical benefits of ensemble methods, setting the stage for a wide range of applications.

Within the healthcare and biomedical signal processing domain, ensemble learning has gained significant traction. Multiple studies have demonstrated that ensembles can improve diagnostic accuracy by fusing diverse feature representations and classification strategies~\cite{rokach2010ensemble, kuncheva2014combining}. In ECG analysis specifically, ensemble approaches have been used to detect arrhythmias~\cite{essa2021ensemble} and other cardiac abnormalities, where each component model targets different ECG characteristics, such as morphological features, frequency-domain information, or temporal patterns. By leveraging specialized expertise from each learner, ensembles can yield higher sensitivity and specificity compared to single-model solutions.

More recently, researchers have begun integrating deep neural networks into ensemble pipelines, exploiting the representational power of deep models while mitigating their tendency to overfit~\cite{kim2020study}. Such ensembles often involve different network architectures, or the same architecture trained under varying hyperparameters or data augmentation schemes. The resulting diversity among base models is a key determinant of ensemble success, ensuring that the final aggregated decision capitalizes on complementary insights.

However, large-scale ensemble solutions based on extensive deep networks can be computationally intensive, especially when models are pretrained on general-domain data rather than specialized ECG tasks. As a result, parameter-efficient fine-tuning strategies, like Low-Rank Adaptation have become increasingly attractive. By selectively updating only a fraction of model parameters, these methods preserve the advantages of ensemble diversity without incurring prohibitive resource costs. In multi-task ECG scenarios, such approaches enable more robust, efficient, and scalable solutions that simultaneously address various cardiac diagnostic objectives.

\section*{Method}

\uline{\textit{Problem Definition}}.
% Jiaying: kind of hard to follow in current writing.
% I remove B for batch, since this is an implementation rather than problem definition. 
% We introduce one ECG waveform, the prediction target should also be one.
One ECG waveform $\mathbf{X}$ can be denoted as $\mathbf{X} \in \mathbb{R}^{C \times T}$, where $C$ is the number of leads, $T$ is the signal length. In our study, we aim to employ a predictive model $f_\theta(\mathbf{X})$ to perform various clinical downstream tasks, thus $\hat{y}=f_\theta(\mathbf{X})$. Specifically, the prediction can be denoted as $\hat{y} \in \mathbb{R}$ for continuous value regression (\textit{e.g.} RR interval estimation and age prediction), $\hat{y} \in \{0, 1\}$ for binary classification (\textit{e.g.} sex classification, potassium abnormality prediction), or $\hat{y} \in \{0, 1, ...,K\}$ for multi-class classification (\textit{e.g.} arrhythmia detection).

%Given one ECG waveform $\mathbf{X} \in \mathbb{R}^{C \times T}$, where $C$ is the number of leads, $T$ is the signal length, various downstream tasks can be addressed, such as $\hat{y}_{RR} \in \mathbb{R}^{B \times 1}$ for RR interval estimation, $\hat{y}_{age} \in \mathbb{R}^{B \times 1}$ for age estimation, $\hat{y}_{sex} \in \{0, 1\}$ for sex classification, $\hat{y}_{Ka} \in \{0, 1\}$ for potassium abnormality prediction, $\hat{y}_{AD} \in \{0, 1, ...,K\}$ for arrhythmia detection, where  $B$ is the batch size and $K$ is the number of classes. 
%\Xiaoda{B is the number of ECG waveforms? }
 
\begin{figure}[htbp]
    \centering
    \includegraphics[width=0.65\textwidth]{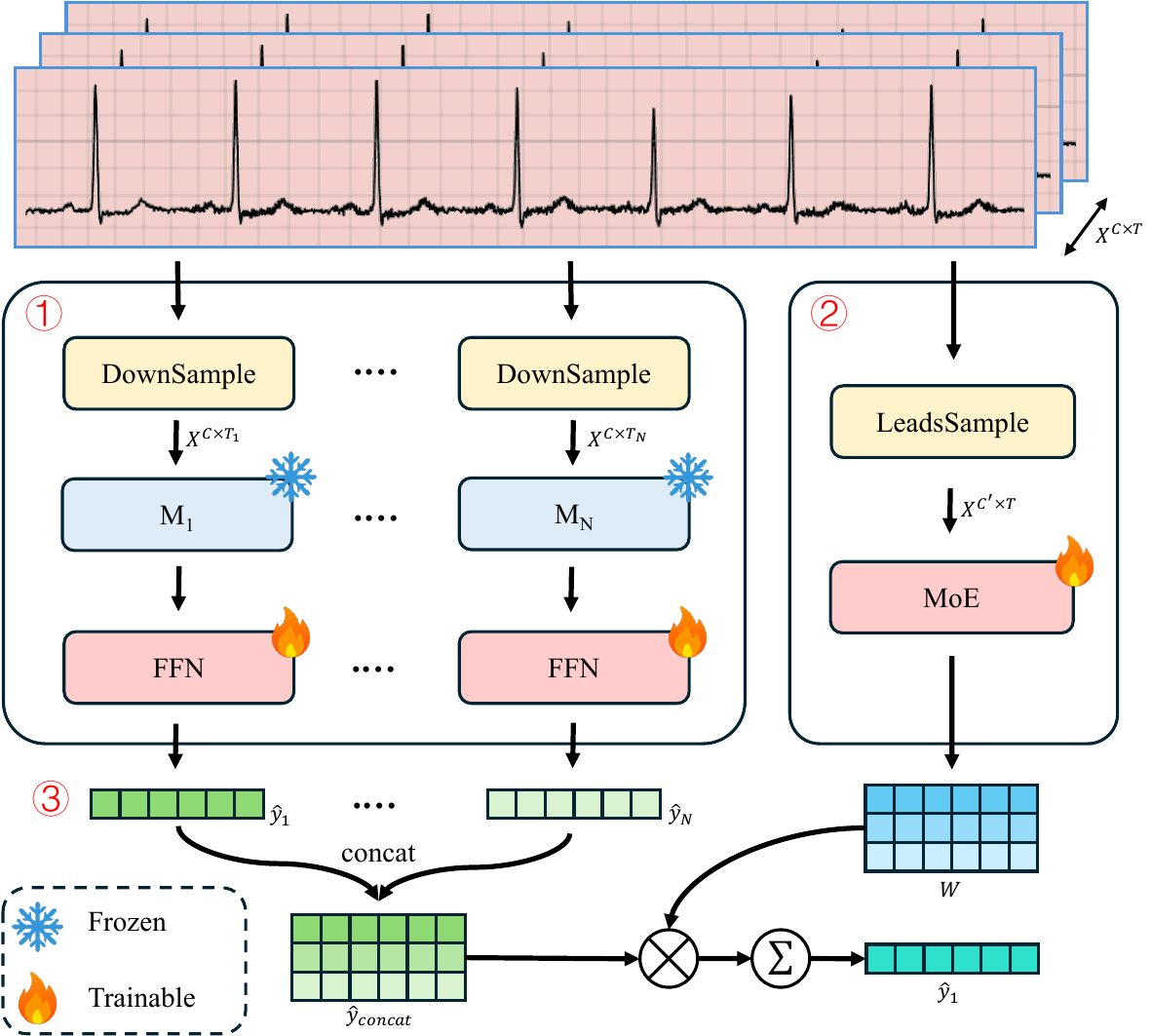}
    \caption{The framework of EnECG. The EnECG framework comprises three main steps. 
\ding{172}~Because each pretrained foundation model $\left(M_1, M_2, \ldots, M_N\right)$ requires a specific input length, we downsample the ECG and feed it into the frozen model. We then add a FFN and fine-tune it to obtain $\left(\hat{y}_1, \hat{y}_2, \ldots, \hat{y}_N\right)$. 
\ding{173}~To reduce training costs, we select a subset of ECG leads and input them into the Mixture of Experts (MoE), which outputs gating probabilities $W$. 
\ding{174}~Finally, we ensemble the results via the weighted sum $\hat{y} = \sum_{i=1}^{N} W_i \hat{y}_i$.}
    \label{model}
\end{figure}

To address the multi-task requirements of ECG analysis while enhancing inference efficiency, we propose a novel framework as shown in Figure~\ref{model}. Our approach fuses outputs from multiple base models $\left(\hat{y}_1, \hat{y}_2, \ldots, \hat{y}_N\right)$ through learnable weighting and employs LoRA for incremental fine-tuning of selected network layers. To determine these weights, we integrate a Mixture of Experts framework, which learns to provide logits-aligned weights after training. Consequently, our model significantly reduces both training and inference overhead while maintaining high predictive accuracy. By unifying regression, classification, and other ECG-related tasks within a single architecture, it strikes an effective balance between performance and computational efficiency.

\uline{\textit{Getting Prediction Logits from Frozen Base Models}}. All base model parameters $\phi$ in our framework are frozen. We then add a feedforward neural network as an output layer at the end of each model and train it. Specifically, for each model, we obtain the logit $\hat{y}_i \in \mathbb{R}$ as: $\hat{y}_i = \mathrm{FFN}\bigl(\mathrm{M}_i(\mathrm{downsample}(X))\bigr)$. Next, we concatenate all logits to form $\hat{y} \in \mathbb{R}^{ N \times L}$, $\hat{y} = \mathrm{concat}(\hat{y}_1, \hat{y}_2, \ldots, \hat{y}_N)$. 

\uline{\textit{Calculating Ensemble Weights}}. After we have the concatenated logits, we use a MoE to learn the weight $W \in \mathbb{R}^{ N \times K}$, $W = \mathrm{MoE}(\mathrm{LeadsSample}(X))$. Finally, we compute the ensemble logit via $\hat{y} = \sum_{i=1}^{N} W_i \hat{y}_i$.

\uline{\textit{LoRA Tuning}}. To mitigate the high resource demands of fine-tuning large-scale pretrained models, we employ Low-Rank Adaptation (LoRA)~\cite{hu2021lora}, which reduces computational cost by updating only a low-rank subset of parameters instead of the full model. In FFN layer and MoE layer, given the initial weight matrix $W_0 \in \mathbb{R}^{d \times k}$ and input $x \in \mathbb{R}^{k}$, the forward pass is: $h = W_0 x$.
Rather than fine-tuning $W_0$, LoRA applies a low-rank update:
\begin{equation}
    \Delta W = BA, \quad A \in \mathbb{R}^{r \times k}, \quad B \in \mathbb{R}^{d \times r}, \quad r \ll \min(d, k),
\end{equation}
% where only $A$ and $B$ are trainable while $W_0$ remains fixed. The adapted weight becomes:
% \begin{equation}
%     W_{\text{adapted}} = W_0 + BA.
% \end{equation}
Thus, the updated forward pass is:
\begin{equation}
    h_{\text{LoRA}} = (W_0 + BA)x.
\end{equation}

By constraining the update to a low-rank space, LoRA maintains the expressiveness of large-scale models while substantially reducing the total number of parameters that must be optimized. This makes fine-tuning both memory- and computation-efficient, enabling us to adapt pretrained models to new tasks with minimal overhead.

\section*{Experiments}

In this section, we evaluate the effectiveness of our proposed model from two key perspectives: performance and efficiency. Specifically, we compare its predictive accuracy across multiple tasks and analyze its computational efficiency in terms of training and inference costs. By examining these aspects, we aim to demonstrate the advantages of our model in achieving a balanced trade-off between accuracy and computational feasibility.

\uline{\textit{Dataset}}. We utilize the MIMIC-IV-ECG~\cite{gow2023mimic} dataset, currently the largest publicly released ECG repository, which contains 800{,}035 diagnostic electrocardiograms acquired from 161{,}352 unique patients. Each recording is 10 seconds in duration, sampled at 500 Hz across 12 leads, resulting in an input representation of dimension $X \in \mathbb{R}^{C \times T}$ with $C=12$ and $T=10 \times 500=5000$.

\uline{\textit{Downstream tasks}}. We evaluate the performance of the benchmark on the following tasks:
\begin{enumerate}
    \item \textbf{\textit{RR Interval Estimation.}} The RR interval, which represents the time between two R-wave peaks in an ECG, is directly calculated from the ECG signal.
    \item \textbf{\textit{Age Estimation.}} Patient age estimation involves analyzing ECG signal characteristics to estimate age, challenging the model to effectively interpret complex signal patterns correlated with physiological aging.
    \item \textbf{\textit{Sex Classification.}} Sex classification is a binary classification task with a balanced ratio of 50\% to 50\%.
    \item \textbf{\textit{Potassium Abnormality Prediction.}} We use ECG strips to predict the Potassium (blood) lab test result which is taken between ECG recording time and one hour after the ECG time. This task is challegning, with imbalanced ratio of 97\% (normal) to 3\% (abnormal).
    \item \textbf{\textit{Arrhythmia Detection.}} We select the 14 most frequently occurring diagnoses, with the remaining ones grouped under ``Others'', resulting in a total of 15 labels.
\end{enumerate}
Among the downstream tasks, RR interval estimation and age estimation are formulated as regression problems, where the prediction target $\hat{y} \in \mathbb{R}$. We use mean absolute error (MAE) as the evaluation metric.

Sex prediction and potassium abnormality prediction are binary classification tasks, with the prediction target $\hat{y} \in \{0, 1\}$. For these tasks, we adopt the F1 score as the evaluation metric.

Arrhythmia detection is treated as a multiclass classification task, where the prediction target $\hat{y} \in \{1, 2, \dots, K\}$, with $K=15$ representing different arrhythmia phenotypes. We use the accuracy to evaluate model performance on it.

\uline{\textit{Baselines}}. We select the following models as baselines and additionally construct an ensemble of them.
\begin{itemize}
    \item TimesNet~\cite{wu2022timesnet}: A deep neural architecture originally introduced for time series multi-tasks such as forecasting, classification, and anomaly detection. It leverages a hierarchical temporal block structure with Fourier-based and convolutional operations to effectively model multi-scale temporal patterns. Its design allows extensions to various downstream tasks, reflecting its versatility across different time series analysis domains.
    \item DLinear~\cite{zeng2023transformers}: A lightweight linear model that decomposes the input signal into trend and seasonal components, then applies separate linear layers to each component. This approach maintains interpretability, significantly reduces computational complexity, and achieves forecasting accuracy comparable to more complex deep learning methods in many scenarios.
    \item MOMENT~\cite{goswami2024moment}: A family of open-source foundation models for time series analysis. It leverages masked pre-training across large-scale, diverse time series datasets and excels in forecasting, classification, anomaly detection, and imputation tasks. By sharing learned representations across different tasks, MOMENT reduces redundant computation and maintains strong generalization, even under limited supervision.
    \item TEMPO~\cite{cao2023tempo}: A specialized generative pre-trained Transformer designed for time series. TEMPO integrates seasonal-trend decomposition with prompt-based training strategies to handle diverse and potentially multi-modal time series inputs. It achieves strong zero-shot and few-shot forecasting performance and provides a flexible multi-task learning framework.
    \item ECG-FM~\cite{mckeen2024ecg}: An open foundation model tailored for ECG data. It applies self-supervised contrastive learning and masked modeling on large-scale ECG corpora, capturing salient features that generalize to downstream classification, regression, and diagnostic tasks. By integrating ECG-domain knowledge into large-scale pre-training, ECG-FM reduces the need for extensive labeled data.
\end{itemize}

\uline{\textit{Research Questions}}. Our experiments are designed to answer following questions.
\begin{itemize}
    \item \textbf{RQ1: Why do we choose the above models and ensemble them?} 
    Numerous time-series foundation models have been proposed in recent years. In this work, we select a representative set of baseline models based on their popularity, diversity in architecture, and applicability to ECG analysis. We further ensemble these baselines to evaluate whether model combination can enhance robustness and performance. All baseline models are evaluated in a zero-shot setting. Each experiment is repeated three times, with 10{,}000 patients randomly selected in each run.
    
    \item \textbf{RQ2: How does the performance of EnECG compare with the baselines?} 
    We propose EnECG to address ECG-based multi-task learning challenges. To assess its effectiveness, we compare its performance with that of the selected fine-tuned baseline models across five downstream tasks. Each experiment is repeated three times, with 10{,}000 patients randomly sampled in each trial and split into training, validation, and test sets following a 70\%{:}20\%{:}10\% ratio.
    
    \item \textbf{RQ3: How does the efficiency of EnECG compare with the baselines?} 
    Given the limited computational resources typically available in clinical settings, model efficiency is a critical factor. We evaluate the efficiency of EnECG by measuring GPU memory consumption and throughput (samples per second), and compare these metrics with those of the baseline models.

    \item \textbf{RQ4: How does the performance of MoE-based ensemble learning compare with other ensemble methods?} In EnECG, we employ a MoE framework to learn expert weights, and then generating a weighted sum of the logits from various foundation models. To evaluate its effectiveness, we conduct comparative experiments against existing ensemble strategies~\cite{lu2023beyond}, including confidence-aware weighting, greedy search, and a sample-aware weight generator for ensemble weight estimation.
\end{itemize}

% \uline{\textit{Implementation Details}} 

\subsection*{Results}

\uline{\textit{Baseline Performance on ECG Downstream Tasks (RQ1)}}. Each baseline model utilizes a different architecture, allowing it to specialize in capturing different ECG features. As detailed in Table~\ref{performance_zs}, none of the baselines performed best on all five ECG downstream tasks. Due to its ECG specific pretraining, ECG-FM generally outperforms other models, underscoring the value of domain, specific knowledge in accurately interpreting ECG signals. However, in the sex classification task, TimesNet achieves superior accuracy, we suppose due to its CNN-based structure that effectively captures localized ECG wave morphology differences between genders in medical diagnostics. Additionally, both DLinear and TEMPO exhibit strong performance in certain specialized tasks, indicating their capacity to model underlying trends and temporal dependencies crucial for specific clinical assessments. Although MOMENT does not achieve top results in the zero-shot scenario, its large-parameter architecture and extensive pretraining on diverse time-series datasets and tasks potentially equip it with generalized feature extraction abilities valuable for more comprehensive ECG interpretation when fine-tuned or incorporated into ensembles. Understanding the individual advantages of these baseline models from a clinical standpoint helps justify combining their strengths and clinically actionable predictions.

\begin{table}[h]
    \caption{Zero-shot performance of baseline models on ECG data. Highlighted are the top \textcolor{teal}{first} and  \textcolor{brown}{second} results. RR Interval Estimation, Age Estimation, Sex Classification, Potassium Abnormality Prediction, Arrhythmia Detection are denoted as RR, Age, Sex, Ka, AD respectively.}
    \centering
    \setlength{\tabcolsep}{3pt}
    \renewcommand{\arraystretch}{1.2}
    \begin{tabular}{lcccccc}
        \toprule
        & & \textbf{TimesNet} & \textbf{DLinear} & \textbf{MOMENT} & \textbf{TEMPO} & \textbf{ECG-FM}  \\
        \midrule
        \multirow{2}{*}{\makecell{\textbf{Regression} \\ \textbf{(MAE)$\downarrow$}}}
        & \textbf{RR}  & $817.0\pm2.5$ & \textcolor{brown}{$816.4\pm2.9$} & $816.6\pm2.1$ & \textcolor{teal}{816.3 $\pm$ 1.9} & \textcolor{teal}{816.3 $\pm$ 1.9}\\ 
        & \textbf{Age} & \textcolor{brown}{$62.28\pm0.36$} & $62.63\pm0.50$ & $62.61\pm0.37$ & $62.33\pm0.38$ & \textcolor{teal}{62.27 $\pm$ 0.38}\\
        \cmidrule(lr){1-7}
        \multirow{2}{*}{\makecell{\textbf{Binary Class} \\ \textbf{(F1)$\uparrow$}}}
        & \textbf{Sex} & \textcolor{teal}{0.60 $\pm$ 0.00} & \textcolor{brown}{$0.51\pm0.08$} & $0.34\pm0.00$ & $0.42\pm0.00$ & $0.33\pm0.00$ \\
        & \textbf{Ka}  & $0.06\pm0.00$ & $0.05\pm0.00$ & $0.02\pm0.00$ & \textcolor{brown}{$0.18\pm0.00$} & \textcolor{teal}{0.35 $\pm$ 0.00} \\
        \cmidrule(lr){1-7}
        {\makecell{\textbf{15 Class} \\ \textbf{(ACC)$\uparrow$}}}
        & \textbf{AD}  & \textcolor{brown}{$0.06\pm0.01$} & $0.03\pm0.02$ & $0.03\pm0.02$ & $0.02\pm0.00$ & \textcolor{teal}{0.07 $\pm$ 0.00} \\
        \bottomrule
    \end{tabular}
    \label{performance_zs}
\end{table}

\uline{\textit{Evaluating the Performance of EnECG vs. Baselines (RQ2)}}. In this experiment, all baseline models were fine-tuned individually on each of the five ECG downstream tasks, with detailed results presented in Table~\ref{performance_comparison}. EnECG consistently achieves superior predictive performance across these tasks, underscoring the effectiveness of integrating specialized foundation models to improve diagnostic accuracy in clinical practice. In the RR interval estimation task, EnECG demonstrates an approximate 38\% improvement over TEMPO, the second-best performing baseline. Such substantial enhancement in RR interval prediction can significantly aid clinicians in better assessing patient cardiac conditions, potentially leading to earlier detection and intervention for arrhythmias or other cardiovascular anomalies.

Although EnECG does not achieve the absolute highest accuracy in the sex classification task, it remains highly competitive, delivering stable predictions that closely match the MOMENT. Importantly, MOMENT requires substantially more computational resources (as detailed in RQ3). It can help hospitals reduce the cost of training high-performance models.

\begin{table}[h]
    \caption{Performance of fine-tuned baseline models on five ECG downstream tasks. Bold values represent the best test performance.}
    \centering
    \setlength{\tabcolsep}{3pt}
    \renewcommand{\arraystretch}{1.2}
    \begin{tabular}{lccccccc}
        \toprule
        & & \textbf{TimesNet} & \textbf{DLinear} & \textbf{MOMENT} & \textbf{TEMPO} & \textbf{ECG-FM} & \textbf{EnECG} \\
        \midrule
        \multirow{2}{*}{\makecell{\textbf{Regression} \\ \textbf{(MAE)$\downarrow$}}}
        & \textbf{RR}  & $304.3\pm4.3$ & $786.0\pm5.4$ & $146.9\pm1.3$ & $141.5\pm2.1$ & $147.3\pm1.3$ & \textbf{87.69 $\pm$ 6.4} \\ 
        & \textbf{Age} & $24.89\pm0.07$ & $28.46\pm0.74$ & $13.41\pm0.45$ & $13.52\pm0.31$ & $13.49\pm0.17$ & \textbf{12.97 $\pm$ 0.61} \\
        \cmidrule(lr){1-8}
        \multirow{2}{*}{\makecell{\textbf{Binary Class} \\ \textbf{(F1)$\uparrow$}}}
        & \textbf{Sex} & $0.51\pm0.05$ & $0.57\pm0.01$ & \textbf{0.69 $\pm$ 0.02} & $0.54\pm0.01$ & $0.52\pm0.05$ & \textbf{0.69 $\pm$ 0.01} \\
        & \textbf{Ka}  & $0.01\pm0.01$ & $0.01\pm0.01$ & $0.49\pm0.00$ & $0.50\pm0.00$ & $0.49\pm0.00$ & \textbf{0.53 $\pm$ 0.01} \\
        \cmidrule(lr){1-8}
        {\makecell{\textbf{15 Class} \\ \textbf{(ACC)$\uparrow$}}}
        & \textbf{AD}  & $0.03\pm0.00$ & $0.48\pm0.02$ & $0.66\pm0.03$ & $0.54\pm0.14$ & $0.49\pm0.03$ & \textbf{0.76 $\pm$ 0.01} \\
        \bottomrule
    \end{tabular}
    \label{performance_comparison}
\end{table}

\begin{figure}[htbp]
    \centering
    \begin{subfigure}[b]{0.45\textwidth}
        \centering
        \includegraphics[width=\textwidth]{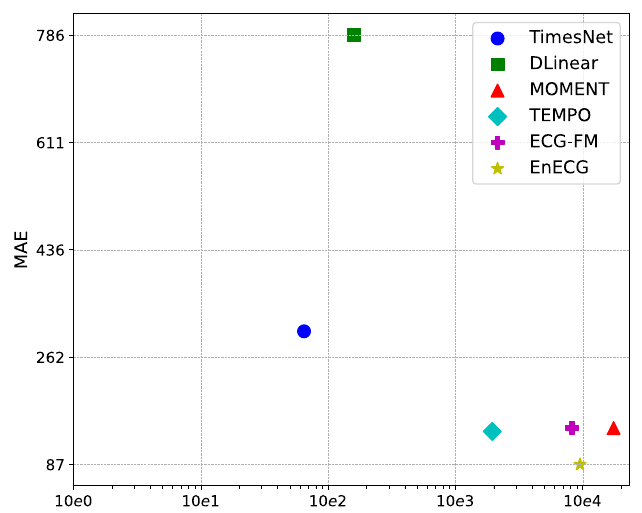}
        \caption{GPU Memory Usage}
        \label{fig:a}
    \end{subfigure}
    \hfill
    \begin{subfigure}[b]{0.45\textwidth}
        \centering
        \includegraphics[width=\textwidth]{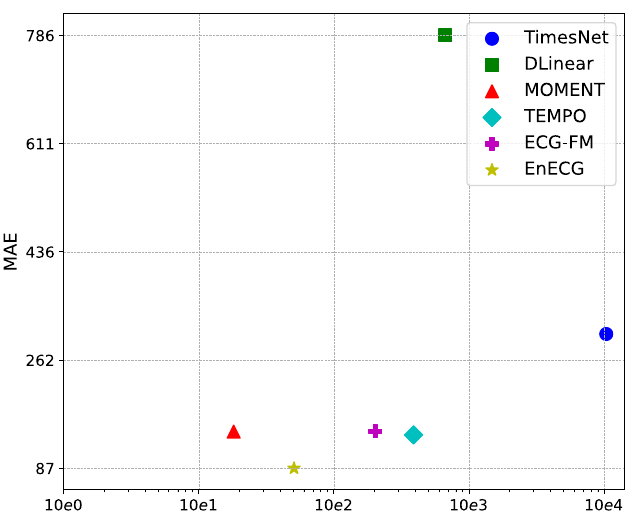}
        \caption{Throughout}
        \label{fig:b}
    \end{subfigure}
    \caption{Training efficiency in the RR interval estimation task. We evaluate both GPU memory usage and throughput alongside model performance. In (a), models closer to the bottom-left corner exhibit better performance with lower memory consumption. In (b), models closer to the bottom-right corner demonstrate better performance and higher training efficiency.}
    \label{efficiency}
\end{figure}

\uline{\textit{Evaluating the Computational Efficiency of EnECG vs. Baselines (RQ3)}}. The efficiency comparison of different models for the RR interval estimation task is illustrated in Figure~\ref{efficiency}. Since all five downstream tasks utilize the same dataset and differ only in their target labels, their computational costs during training are expected to be similar. We evaluate efficiency by measuring active GPU memory usage (in MB) and throughput (in iterations per second).

As shown in Figure~\ref{efficiency}(a), EnECG achieves the highest performance while maintaining GPU memory usage below 10 GB. Such modest computational requirements are achievable with consumer-grade GPUs, significantly reducing hardware expenses and infrastructure costs in hospital settings, thus making advanced AI-assisted ECG interpretation accessible even for facilities with limited resources. Although TimesNet and DLinear demonstrate lower GPU memory consumption, their substantially inferior predictive accuracy limits their practical utility, especially in critical healthcare scenarios where accuracy directly impacts patient safety and outcomes. On the other hand, MOMENT, TEMPO, and ECG-FM utilize a comparable level of GPU resources, highlighting EnECG’s superior resource efficiency and effectiveness in clinical environments.

From Figure~\ref{efficiency}(b), EnECG achieves a throughout exceeding 10 samples per second, meaning a single patient's ECG data can be analyzed and interpreted within approximately 0.1 seconds. Such quick processing time is particularly significant in medical contexts, where timely diagnostic decisions can help improve emergency responsiveness. Rapid and accurate ECG analysis supports prompt medical interventions, minimizes potential delays in diagnosis, and contributes directly to improved patient outcomes. Consequently, EnECG’s combination of high computational efficiency and strong predictive performance makes it especially suited to real-world medical applications.

\uline{\textit{Evaluating the Ensemble Weighting of EnECG vs. Existing Methods (RQ4)}}. Since the zero-shot ensemble method relies on computing the cosine similarity between prediction logits and ground truth labels, it is inherently less suitable for regression tasks, which require continuous numerical outputs critical for clinical measurements such as RR intervals or age estimation. As shown in Table~\ref{ensemble_comparison}, EnECG consistently outperforms alternative ensemble strategies, including the training-free and tuning-based methods, across four downstream tasks. 

Moreover, EnECG demonstrates notably lower standard deviations in predictive performance, reflecting greater model stability, particularly in regression tasks. This stability is medically significant, as variability or inconsistencies in predictive outputs can lead to clinical misinterpretations or delayed decision-making, potentially compromising patient safety. Reliable predictions are crucial in clinical environments to guide timely interventions and optimize patient outcomes. By effectively integrating multiple specialized foundation models through a MoE framework, EnECG maximizes the strengths and mitigates individual model limitations, resulting in more consistent, dependable clinical predictions.

\begin{table}[h]
    \caption{Performance comparison of ensemble weighting methods. We compare three approaches: confidence-aware weighting (``Zero-shot''), greedy search (``Training-free''), and a sample-aware weight generator (``Tuning''), as defined in the original paper~\cite{lu2023beyond}. Bold values denote the best test performance.``-'' means can not applied in regression task.}
    \centering
    \setlength{\tabcolsep}{3pt}
    \renewcommand{\arraystretch}{1.2}
    \begin{tabular}{lccccc}
        \toprule
        & & \textbf{Zero-shot} & \textbf{Training-free} & \textbf{Tuning} & \textbf{EnECG} \\
        \midrule
        \multirow{2}{*}{\makecell{\textbf{Regression} \\ \textbf{(MAE)$\downarrow$}}}
        & \textbf{RR}  & - & $369.7\pm317.8$ & $140.6\pm48.7$ & \textbf{87.69 $\pm$ 6.4}  \\ 
        & \textbf{Age} & - & $26.43\pm21.06$ & $13.07\pm1.03$ & \textbf{12.97 $\pm$ 0.61} \\
        \cmidrule(lr){1-6}
        \multirow{2}{*}{\makecell{\textbf{Binary Class} \\ \textbf{(F1)$\uparrow$}}}
        & \textbf{Sex} & $0.60\pm0.05$ & $0.34\pm0.00$ & $0.57 \pm 0.01$ & \textbf{0.69 $\pm$ 0.01} \\
        & \textbf{Ka}  & $0.49\pm0.00$ & $0.49\pm0.00$ & $0.49\pm0.00$ & \textbf{0.53 $\pm$ 0.01} \\
        \cmidrule(lr){1-6}
        {\makecell{\textbf{15 Class} \\ \textbf{(ACC)$\uparrow$}}}
        & \textbf{AD}  & $0.21\pm0.03$ & $0.12\pm0.00$ & \textbf{0.78 $\pm$ 0.02} & 0.76 $\pm$ 0.01 \\
        \bottomrule
    \end{tabular}
    \label{ensemble_comparison}
\end{table}

\subsection*{Conclusion}

In this study, we proposed EnECG, a Mixture-of-Experts (MoE)-based ensemble framework designed for multi-task electrocardiogram (ECG) analysis. By integrating multiple specialized models, EnECG consistently achieves superior accuracy and stability across clinically important ECG tasks, including regression (RR interval and age estimation) and classification (sex, potassium abnormality and arrhythmia). Our work also demonstrates efficient computational resource usage. This efficiency facilitates rapid and reliable clinical decision-making, potentially enhancing patient outcomes through timely diagnostics. To address the limitation of EnECG (e.g. outperforming all baselines except Tuning ensemble in arrhythmia detection), future work will focus on adaptive ensemble strategies and dynamic expert selection mechanisms to further enhance predictive accuracy across individual ECG tasks, as well as develop a comprehensive ECG foundation model system designed to further support clinical decisions. 

\subsection*{Acknowledgement}

This research was partially supported by the US National Science Foundation under Award Numbers 2319449, 2312502, and 2442172, as well as the US National Institute of Diabetes and Digestive and Kidney Diseases of the US National Institutes of Health under Award Number K25DK135913. We thank for the computing resources provided by the iTiger GPU cluster~\cite{sharif2025cultivating} supported by the NSF MRI program.

% References as numbers
\makeatletter
\renewcommand{\@biblabel}[1]{\hfill #1.}
\makeatother

% unstr is used to keep citation order
\bibliographystyle{vancouver}
\bibliography{amia}  

\end{document}